# Jointly Contrastive Representation Learning on Road Network and Trajectory

Zhenyu Mao
SenseTime Research
Shanghai, China
maozhenyu@sensetime.com

Ziyue Li*
University of Cologne
50923 Cologne, NRW, Germany
EWI gGmbH
50827 Cologne, NRW, Germany
zlibn@wiso.uni-koeln.de

Dedong Li
SenseTime Research
Shanghai, China
lidedong1@sensetime.com

Lei Bai
The Shanghai AI Laboratory
Shanghai, China
baisanshi@gmail.com

Rui Zhao
SenseTime Research
Qing Yuan Research Institute of
Shanghai Jiao Tong University
Shanghai, China
zhaorui@sensetime.com

## ABSTRACT

Road network and trajectory representation learning are essential for traffic systems since the learned representation can be directly used in various downstream tasks (*e.g.*, traffic speed inference, travel time estimation). However, most existing methods only contrast within the same scale, *i.e.*, treating road network and trajectory separately, which ignores valuable inter-relations. In this paper, we aim to propose a unified framework that jointly learns the road network and trajectory representations end-to-end. We design domain-specific augmentations for road-road contrast and trajectory-trajectory contrast separately, *i.e.*, road segment with its contextual neighbors and trajectory with its detour replaced and dropped alternatives, respectively. On top of that, we further introduce the road-trajectory cross-scale contrast to bridge the two scales by maximizing the total mutual information. Unlike the existing cross-scale contrastive learning methods on graphs that only contrast a graph and its belonging nodes, the contrast between road segment and trajectory is elaborately tailored via novel positive sampling and adaptive weighting strategies. We conduct prudent experiments based on two real-world datasets with four downstream tasks, demonstrating improved performance and effectiveness.

## 1 INTRODUCTION

*Road network* and *trajectory* are two essential components in spatiotemporal urban transportation systems [9, 29]: Road network is typical graph data, which contains the topological and auxiliary contextual information between the road segments [2, 16, 41]; Trajectory instead is sequence data with consecutive road segments, which is spatiotemporal dynamic and includes the mobility semantics [24, 37, 42–44].

As shown in Figure 1, the static road network and the dynamic trajectory give different and complementary insights about two road segments, *i.e.*, A and B. The road network indicates A and B are distant since their shortest path (plotted in black-dotted line) along the network is rather long, which may result in different embeddings for A and B; The dynamic trajectory yet indicates the travel between A and B is quite frequent via another route (plotted in purple line), which indicates strong spatial correlations and may lead to similar embeddings for A and B. Therefore, only using either road network or trajectory may only capture partial information.

The existing representation learning studies for road network and trajectory have been constantly conducted in parallel and separately. Specifically, road network representation embeds the node-like, local-scale road segment into a representation vector containing the graph topology [15], hierarchical information [38], and intersection [35] structure within the road network. Trajectory representation embeds the sequence-like, global-scale trajectory using the sequential context within the trajectory [19]. However, those separated treatments ignore the valuable inter-relational information between road network and trajectory.

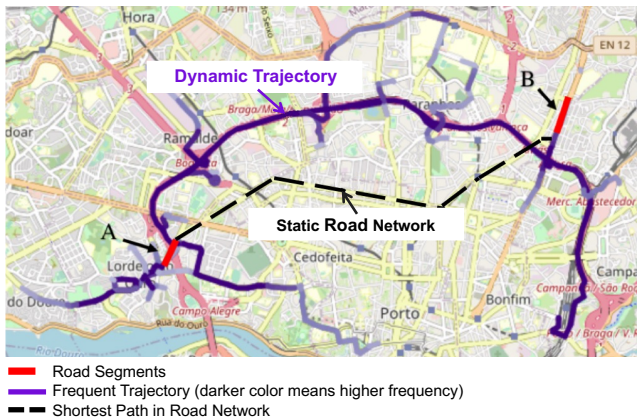

**Figure 1: An example of how road network and trajectory offer different and complementary information to representation learning. Static road network shows the shortest path between segment A and B via black-dotted line; Dynamic trajectory shows that the most popular route to travel from A to B is yet via the purple solid line and a darker line means a higher travel frequency.**

In the recent, the state-of-the-art methods have proved that bringing the road network representation as semantics into trajectory representation learning [12], or oppositely incorporating trajectory representation as contexts into road network representation [6] offers better representations than the methods ignoring the inter-relations between the road network and trajectory [25]. The inter-relations across the road-trajectory pair are two-fold, 1) road network topology constrains the mobility of trajectories, and 2) the trajectory's traveling patterns reflect the semantic dependencies between road segments [44]. To integrate the inter-relations, these methods [6, 12, 25, 44] combine road network and trajectory in a two-step manner, i.e., first learn representation for one scale and then use it as a cornerstone to learn another scale's representation. Though effective, the drawbacks of these methods are two-fold: Firstly, it would suffer from error propagation; Secondly, each step's objective is to predict similar pairs within the same scale, i.e., road-road and trajectory-trajectory pairs, which reduces to within-scale learning without building the bridge between them.

To fill the gap, this work aims to better capture the knowledge across the road network-trajectory pair on top of preserving within-road network and within-trajectory information.

Contrastive learning [5, 28] has proved its advantage in graph self-supervised learning [26], especially for cross-scale representation, by capturing the complex dependencies between different scales' representations with a loss to maximize their mutual information (MI) [1]. However, under the umbrella of node-graph contrast [45], a positive pair is usually strictly defined by whether it is on the same graph of the node. Nevertheless, the connection between road segment and trajectory can be more flexible and complex. Several recent studies have applied contrastive learning to path representation [40] and navigation [25], but their works still limited to within-scale and ignore the whole road network.

Based on the foundation laid by existing road segment and trajectory representation learning and node-graph contrastive learning, we develop a unified framework for road-road, trajectory-trajectory, and road-trajectory contrast, which is more elaborately tailored for the specific and significant domain of transportation. Precisely, we design a contrastive learning framework to simultaneously maximize the total mutual information from both within-scale (road-road, trajectory-trajectory) and cross-scale (road-trajectory) contrasts. Instead of simply stacking up three contrasts together, this unified framework carefully designs each contrast respecting the intrinsic properties of road segment and trajectory: (1) for road-road contrast, augmentation based on the contextual graph with structure and transition views guides the search for positive pairs; (2) for trajectory-trajectory contrast, trajectory-related augmentation based on detour, replace, and drop-out is designed to generate positive samples; (3) furthermore, road-trajectory cross-scale contrast loss is innovatively introduced and tailored for road segment and trajectory, which we will describe below in detail.

Unlike general graph contrastive learning that contrasts a graph with its belonging nodes [1, 13], the relation between road network and trajectory is more complex. For example, a road segment $s$ that is not necessarily on the trajectory $\tau$ can still be regarded as a positive pair when this road segment is well connected with the trajectory $\tau$ by another route. Thus, we design a metric, i.e., RoadSegment-Trajectory (RS-T) distance, to guide the search of positive pairs in a more flexible way. RS-T distance measures the relevance between road segments and trajectories by considering alternative routes. Positive road-trajectory pairs can adaptively contribute to the contrastive loss according to their RS-T distance.

Our contributions are summarized as below:

- We propose a unified and end-to-end model that jointly learns road network and trajectory representations by maximizing the mutual information with traffic domain's tailored road-road, trajectory-trajectory, and road-trajectory losses.
- Specifically for the road-trajectory cross-scale contrast, we define a novel distance for road-trajectory and design a loss with adaptive weights for positive pairs.
- We conduct prudent experiments based on two public datasets in four traffic-related tasks and prove our method's improved performance.

The rest of the paper is organized as follows: firstly, we will review related works from the perspective of application and methodology in Section 2; Then, we will give the notation and preliminary in Section 3. We propose the method in Section 4, and we apply the proposed method in two car-hailing trip datasets with various downstream tasks in Section 5. Lastly, we give our conclusion and future work in Section 6.

## 2 RELATED WORK

In this section, we mainly review two types of related works: (1) application-oriented works that try to combine road and trajectory representation learning, and (2) methodology-oriented works that aim at graph contrastive learning. The limitation of (1) is that: they are still conducted in a two-step manner, with the drawback of error propagation and the constraints in within-scale learning. The limitation of (2) is that: the cross-scale of a graph is node-graph, which strictly defines a positive pair by whether it is on the same graph of the node. We will give detailed reviews below.

## 2.1 Representation on road network and trajectory

Previous works on road networks and trajectories are mainly designed for specific tasks, such as traffic forecasting [8, 23] and travel time estimation [18, 20]. Nevertheless, learning more general representations is more valuable. As mentioned before, representations were learned from graph structure and sequence information of road networks and trajectories, respectively and separately [15, 33].

The latest studies [6, 12, 35, 38, 40] have proved that combining road network and trajectory can yield better representation than learning one scale alone. IRN2Vec [35] explores the intrinsic geospatial properties of nodes and their relationships in road networks. HRNR model [38] constructs a hierarchical neural architecture to learn embeddings from different levels. GTS [12] brings the road network's POI information into trajectory representation, and Toast [6] oppositely brings trajectory (generated by Skip-gram) as context to road network representation. However, as mentioned before, the methods above are still conducted in a two-step manner where road network and trajectory are learned separately and add one as extra knowledge to the other; thus, it has two main drawbacks: firstly, it suffers error propagation; secondly, each step still constraints in within-scale learning, therefore, such paradigms cause information loss, and representation learned in the second step can not conversely help the first step.

One of the most recent and related works is Path InfoMax (PIM) [40]. It defines negative samples within-trajectory contrast with a range from non-overlapping paths to mostly-overlapping paths. However, this work is still limited to within the scale of trajectory only and ignores the whole road network. We aim to better connect the road network and trajectory via cross-scale contrast. Specifically, we try to build an end-to-end framework to jointly model road network and trajectory via unified contrasts for road-road, trajectory-trajectory, and road-trajectory.

## 2.2 Contrastive Representation Learning

Contrastive methods measure the loss in latent space by contrasting samples from a distribution that contains dependencies of interest and distribution that does not [4]. Many state-of-the-art methods construct contrasting pairs in a multi-scale manner. For example, Deep InfoMax (DIM) [14] scores the MI between pixel-level and global features of images. Deep Graph Infomax (DGI) [33] contrasts between nodes and high-level summaries of graphs to learn graph representation. Graphical Mutual Information (GMI) [27] proposes to learn node and edge representations in graphs by contrasting the input and output of graph neural networks (GNN).

To better construct a cross-contrast, it is worth mentioning that an effective and reasonable augmentation is essential such that contrasting congruent and in-congruent views could encourage the encoders to learn rich representations. Unlike image data, the augmentation of graph data is challenging because of its non-euclidean structure. Various augmentation schemes have been proposed, including removing edges, masking features [45, 46], and diffusion transformation [13]. They demonstrated that contrasting node and graph encoding across different views could achieve better results. However, the works above still stay inside the box of the node-graph contrast, and their augmentation methods do not consider the nature of road networks and trajectories, which might be too ideal or unrealistic for practical applications, e.g., transportation.

The most recent and related work has succeeded in applying contrastive learning to the navigation domain [25]. They conduct the contrast within instruction (language) and within vision, which utilize sequence-based and image-based augmentations, respectively. However, it still remains within-scale contrast. Moreover, domain difference also plays a significant role. Previous contrastive learning applied on path representation [40] falls in the same category, they define negative samples within-trajectory contrast with a range from non- overlapping paths to mostly-overlapping paths.

In our case, we focus on building bridges between two types of data, i.e., road network and trajectory, which could be equivalent to graph and sequence data, respectively. Moreover, there are complex spatial and semantic connections between the two. Encoding their information simultaneously is not trivial. To solve this, we propose to learn road network and trajectory representations jointly in a unified model by contrasting road-road, trajectory-trajectory, and road-trajectory, with each augmentation and contrast delicately designed for this domain.

## 3 NOTATION AND PRELIMINARIES

In this section, we give the notation, preliminaries, and formal problem definition. Scalar is denoted in italics, e.g., $n$; vector by lowercase letter in boldface, e.g., $\mathbf{h}$; matrix by uppercase boldface letter, e.g., $\mathbf{A}$; a set by script capital, e.g. $\mathcal{G}$.

*Definition 3.1 (Road Network).* A road network is a directed graph $\mathcal{G} = <\mathcal{S}, \mathbf{A}_{\mathcal{S}}>$, where $\mathcal{S}$ is a vertex set of $|\mathcal{S}|$ road segments, and $\mathbf{A}_{\mathcal{S}} \in \mathbb{R}^{|\mathcal{S}| \times |\mathcal{S}|}$ is the adjacency matrix. An entry $\mathbf{A}_{\mathcal{S}}[s_i, s_j]$ is a binary value indicating whether the end of $s_i$ and the start of $s_j$ share a common intersection. There are works [35] defining the intersections of road network as vertices, but we define road segments as vertices in the widely adopted way.

*Definition 3.2 (Trajectory).* A trajectory $\tau$ is a sequence that consists of $|\tau|$ consecutive road segments, denoted as $\tau = \{s_{k_1}, s_{k_2}, ..., s_{k_n}\}$. Trajectory captures the movement of an object in the road network $\mathcal{G}$.

*Definition 3.3 (Road Network and Trajectory Representation Learning).* Given a road network $\mathcal{G} = <\mathcal{S}, \mathbf{A}_{\mathcal{S}}>$ and a set of historical trajectories $\mathcal{T}$, we aim to learn a representation matrix for the road network $\mathbf{H}_{\mathcal{S}} \in \mathbb{R}^{|\mathcal{S}| \times d}$, where the $s_i$th row $\mathbf{h}_{s_i}$ is the embedding of road segment $s_i$, and a representation vector $\mathbf{h}_{\tau} \in \mathbb{R}^d$ for each trajectory $\tau \in \mathcal{T}$.

## 4 PROPOSED METHOD

We will present the proposed jointly contrastive learning method for the road network and trajectory representations. We first introduce road network and trajectory encoding. Then we define the three components of the loss function, which estimate the MI between road-road, trajectory-trajectory, and road-trajectory pairs, respectively. Finally, the representations can be obtained by simultaneously maximizing all three MI estimators. The proposed framework is shown in Figure 2.

## 4.1 Encoding Module

The encoding module aims at transferring road networks and trajectories into dense vectors in latent space. This module mainly consists of a graph encoder for road network representation and a sequence encoder for trajectory representation. We use Graph Attention Networks (GATs) [32] and Transformer Encoder [31] for graph and sequence encoder, respectively.

### 4.1.1 Graph Encoder for Road Segment Representation. Since road networks are directed graphs, spectral-domain methods are not applicable. We encode road segment representations with GAT:

$$\mathbf{H}_S = \text{GAT}(\mathbf{V}_S, \mathbf{A}_S) \tag{1}$$

The input $\mathbf{V}_S \in \mathbb{R}^{|S| \times d}$ is the initial embedded vectors of road segments, and $\mathbf{A}_S$ is the adjacency matrix for a given graph. In our initial training, we choose structure graph $\mathbf{A}_S$ to capture the network topology, which will be defined in detail in Section 4.2.1. Given the graph encoder $\text{GAT}(\cdot) : \mathbb{R}^{|S| \times d} \times \mathbb{R}^{|S| \times |S|} \mapsto \mathbb{R}^{|S| \times d}$, the representation of road segment $\mathbf{H}_S \in \mathbb{R}^{|S| \times d}$ can be learned.

GAT is chosen since it is a well-developed graph neural network, and suitable for directed graph. It has been applied in several conventional studies [21, 34, 38], and proved its superior performance. We give the formulation of GAT as follows:

$$\text{GAT}(\mathbf{V}_S, \mathbf{A}_S) = \|_{h=1}^{H} \sigma\left(\sum_{j \in N_i} \alpha_{ij}^h \mathbf{W}^h \mathbf{v}_j\right), \tag{2}$$

$$\alpha_{ij} = \frac{exp(\sigma(\mathbf{a}^\top[\mathbf{W}\mathbf{v}_i \| \mathbf{W}\mathbf{v}_j]))}{\sum_{k \in N_i} exp(\sigma(\mathbf{a}^\top[\mathbf{W}\mathbf{v}_i \| \mathbf{W}\mathbf{v}_k]))}, \tag{3}$$

where $\mathbf{v}_i$ is one of road segment vectors in $\mathbf{V}_S$, the attention has $H$ heads, $\|$ represents concatenation, $\sigma$ is the activation function, $N_i$ is neighbor set of road segment $s_i$ according to the graph adjacency matrix $\mathbf{A}_S$, $\mathbf{W}^h$ is the weight matrix of the $h$th head, $\mathbf{W}$ is a shared weight matrix, $\mathbf{a}$ is a weight vector.

### 4.1.2 Sequence Encoder for Trajectory Representation. We select road segments contained in each trajectory and encode them as sequences to obtain trajectory representation. Specifically, given a trajectory $\tau = \{s_{k_1}, s_{k_2}, ..., s_{k_n}\}$, we stack the sequence encoder on the top of graph encoder, the input of sequence encoder is $\mathbf{H}_\tau = \{\mathbf{h}_{s_{k_1}}, \mathbf{h}_{s_{k_2}}, ..., \mathbf{h}_{s_{k_n}}\}$, where $\mathbf{h}_{s_i} \in \mathbb{R}^d$ is the $s_i$-th row of $\mathbf{H}_S$, as well as the positional embedding $\mathbf{P} \in \mathbb{R}^{|\tau| \times d}$. We use sinusoidal positional encoding whose calculation is formulated by:

$$\mathbf{P}[p, 2i] = \sin(p/\omega), \mathbf{P}[p, 2i+1] = \cos(p/\omega) \tag{4}$$

where $p$ is the position index, and $\omega = 1/10000^{2i/d}$.

To learn a summarized trajectory representation $\mathbf{h}_\tau \in \mathbb{R}^d$, we apply transformer encoder $\text{TransEnc}(\cdot) : \mathbb{R}^{|\tau| \times d} \mapsto \mathbb{R}^{|\tau| \times d}$ followed by a mean-pooling operation $\text{Pool}(\cdot) : \mathbb{R}^{|\tau| \times d} \mapsto \mathbb{R}^d$ on the input sequences. The formulation is given as follows:

$$\mathbf{h}_\tau = \text{Pool}(\text{TransEnc}(\mathbf{H}_\tau + \mathbf{P})) \tag{5}$$

Transformer layers encode sequence data by weighting the significance of each part of the inputs. The architecture adopts the mechanism of multi-head self-attention, which is defined below:

$$\text{Attention}(\mathbf{Q}, \mathbf{K}, \mathbf{V}) = \text{softmax}\left(\frac{\mathbf{Q}\mathbf{K}}{\sqrt{d_k}}\right)\mathbf{V} \tag{6}$$

$$head_i = \text{Attention}(\mathbf{X}\mathbf{W}_i^Q, \mathbf{X}\mathbf{W}_i^K, \mathbf{X}\mathbf{W}_i^V) \tag{7}$$

$$\text{MultiHead}(\mathbf{X}) = \|_{i=1}^{h} head_i \cdot \mathbf{W}^O, \tag{8}$$

where $\mathbf{W}_i^Q, \mathbf{W}_i^K, \mathbf{W}_i^V \in \mathbb{R}^{d \times d_k}$ are weight parameters for query, key and value vectors that project input representation $\mathbf{X}$ into $h$ attention heads, $d_k = d/h$, $\|$ is the concatenation operation that combines multi-head attention outputs, and $\mathbf{W}^O \in \mathbb{R}^{d \times d}$ is the output weight.

The output representations are then fed into a two-layer feedforward network. Formally, we have the formulation as follows:

$$\text{FFN}(\mathbf{X}) = \text{ReLU}(\mathbf{X}\mathbf{W}_1 + \mathbf{b}_1)\mathbf{W}_2 + \mathbf{b}_2 \tag{9}$$

where $\mathbf{W}_1, \mathbf{W}_2$ and $\mathbf{b}_1, \mathbf{b}_2$ are weight and bias parameters in dense layer, and $\text{ReLU}(x) = \max(0, x)$ [10] is the activation function.

To alleviate possible training difficulty caused by the increasing depth of neural network, residual connection and layer normalization are applied on the two sub-layers. The formulation of Transformer encoder layer is given as follows:

$$\mathbf{Z} = \text{LayerNorm}(\mathbf{X} + \text{MultiHead}(\mathbf{X})) \tag{10}$$

$$\text{TransEncLayer}(\mathbf{X}) = \text{LayerNorm}(\mathbf{Z} + \text{FFN}(\mathbf{Z})) \tag{11}$$

The outputs of GAT and Transformer Encoder, i.e., $\mathbf{H}_S$ and $\mathbf{h}_\tau$, are taken as the final representations of the road network and trajectory, respectively.

## 4.2 Contrastive Loss Function

Contrastive loss measures the agreements among samples in latent space to gather samples from a distribution that contains dependencies of interest (i.e., positive samples) together, whereas pushing negative samples away [4]. In order to train the framework end-to-end, we define three components of the loss function with MI estimator $\mathcal{I}$ to score the agreements between road-road, trajectory-trajectory, and road-trajectory pairs, respectively. Throughout the paper's augmentation scheme, we will introduce how to select positive samples for the three contrasts, and the rest samples are automatically deemed negative samples.

### 4.2.1 Road-Road Contrastive Loss. Unlike the general node-node contrast, we emphasize that road segments are nodes in a road network with abundant contextual meanings. The road-road loss scores the agreement between each road segment and its context:

$$\mathcal{L}_{SS} = -\frac{1}{|S|} \sum_{s_i \in S} \left[ \frac{1}{|C(s_i)|} \sum_{s_j \in C(s_i)} \mathcal{I}(\mathbf{h}_{s_i}, \mathbf{h}_{s_j}) \right] \tag{12}$$

where $C(s_i)$ is the context set of $s_i$ and defined as contextual graph. $C(s_i)$ will consider both structural and transisitional neighbors of $s_i$, denoting as $\mathbf{A}_S$ and $\mathbf{A}_T$, respectively [22, 38].

For the structure view, we take the direct neighbors of $s_i$ in the road network graph as its structural context, which is recorded in the graph adjacency matrix $\mathbf{A}_S$. $\mathbf{A}_S[s_i, s_j] = 1$ when $s_i, s_j$ are connected, otherwise it is 0.

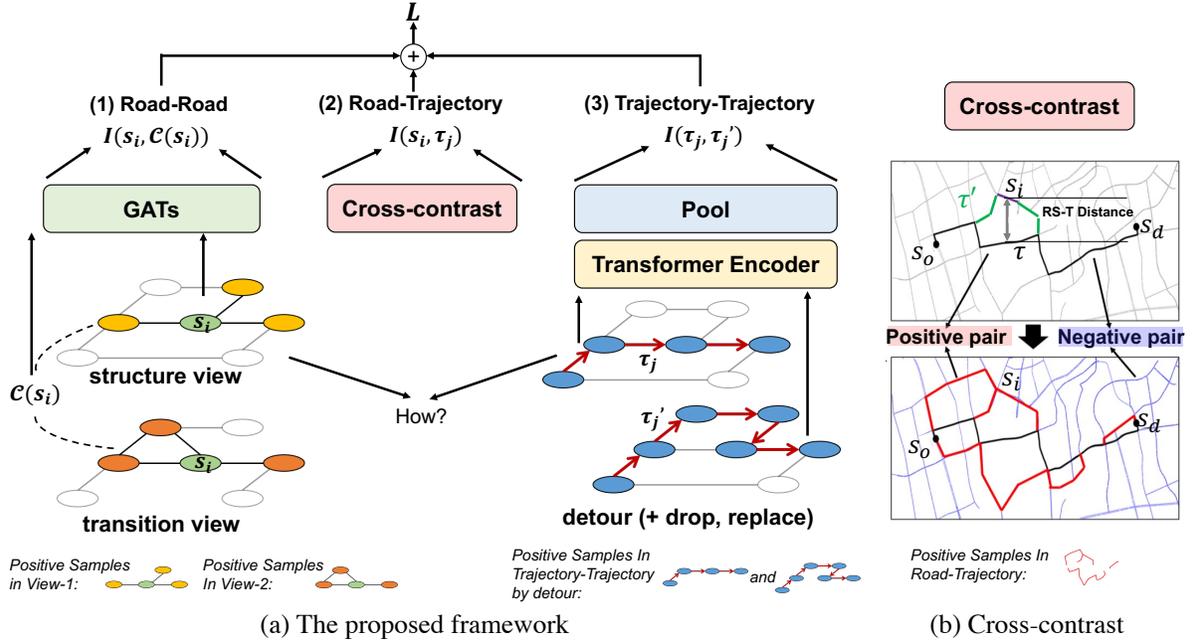

(a) The proposed framework                                            (b) Cross-contrast

Figure 2: (a) The proposed framework for jointly contrastive representation learning on road network and trajectory, with domain-specific augmentations on (1) road-trajectory within-scale contrast, (2) road-trajectory cross-scale contrast, and (3) trajectory-trajectory within-scale contrast. The question remains "how" to conduct the road-trajectory cross-scale contrast. (b) RoadSegment-Trajectory distance considering the potential route options is designed for road-trajectory cross-scale contrast.

For the transition view, we calculate the transition matrix $A_T \in \mathbb{R}^{|\mathcal{S}| \times |\mathcal{S}|}$ by collecting frequently occurred road pairs in trajectory data. Formally, given a trajectory set $\mathcal{T} = \{\tau_1, \tau_2, ..., \tau_n\}$, we initialize the entry of matrix $M_T[s_i, s_j]$ with the frequency that $s_i$ has reached $s_j$ in all trajectory sequences. Then we apply normalization along the row axis, followed by a binarization operation with a threshold $t$ to obtain the transition matrix $A_T$.

$$M_T[s_i, s_j] = \text{Count}(\{\tau_k \in \mathcal{T} | s_i \to s_j \text{ in } \tau_k\}), \qquad (13)$$

$$A_T[s_i, s_j] = \begin{cases} 1, & \text{Norm}(M_T)[s_i, s_j] \geq t; \\ 0, & \text{otherwise.} \end{cases} \qquad (14)$$

$A_T$ is the adjacency matrix of the transition view, and the direct neighbors of $s_i$ in this view is the transitional context.

In conclusion, as shown in Figure. 2. (a)-(1), for road-road contrast, we defined the *context*, *i.e.*, structure and transition neighbors from $A_S$ and $A_T$, as positive samples for road segments, then those non-neighbors are negative samples.

#### 4.2.2 Trajectory-Trajectory Contrastive Loss.
Conventional works [6, 19] learn trajectory representation with denoising methods by predicting the original trajectory with noisy input. We reformulate the loss in a contrastive way:

$$\mathcal{L}_{TT} = -\frac{1}{|\mathcal{T}|} \sum_{\tau_i \in \mathcal{T}} \mathcal{I}(h_{\tau_i'}, h_{\tau_i}) \qquad (15)$$

where $\tau'$ and $\tau$ are the corresponding noisy positive input and the original trajectory.

To generate the noisy positive trajectory sequence, except the traditional augmentation methods, such as random masking and replacement, we also use a trajectory-specific augmentation method, namely detour. Intuitively, an alternative route with the exact origin and destination can also be regarded as a positive pair. Thus, we randomly select a short part of a trajectory and replace it with another path with the same origin and destination to generate a detour positive pair, as demonstrated in Figure. 2. (a)-(3). Besides, generic augmentations such as random masking and replacement are also applied. Other different raw trajectories are negative samples.

#### 4.2.3 Road-Trajectory Contrastive Loss.
If following the node-graph contrast, the road-trajectory loss can be straightforwardly defined by road segment $\mathcal{S}$ and trajectory $\mathcal{T}$:

$$\mathcal{L}_{ST} = -\frac{1}{|\mathcal{T}|} \sum_{\tau_j \in \mathcal{T}} \left[ \frac{1}{|\tau_j|} \sum_{s_i \in \tau_j} \mathcal{I}(h_{s_i}, h_{\tau_j}) \right] \qquad (16)$$

$\mathcal{L}_{ST}$ scores the agreement between road representation and trajectory representation. Thus, road segments related to the trajectory can get close in representation space, and trajectories made up of similar road sequences will render similar representations.

However, Eq. (16) defines positive samples strictly as the road segments on the exact trajectory. In contrast, the road segments that are not necessarily on the trajectory yet quite adjacent to it can still be considered positive samples. The reason is that, although the origin and destination of a trajectory are irreplaceable, in reality, the route choice can be flexible with alternatives: those "off-the-track" road segments yet on the alternatives of the trajectory should be considered as positive samples. To this end, we design a metric called RoadSegment-Trajectory (RS-T) distance to define positive samples more reasonably and flexibly. Precisely, as shown in Figure 2. (b), the origin and destination (OD) of trajectory $\tau$ (in color black) are denoted as $s_o$ and $s_d$. For an arbitrary road segment $s_i$, we generate a route $\tau'$ (in color green) that shares the same OD ($s_o$ and $s_d$) as $\tau$, and travels from $s_o$ to $s_i$ with shortest path, then from $s_i$ to $s_d$ also with shortest path. We define RS-T distance $\mathbf{w}_\tau$ as follows:

$$\mathbf{w}_\tau[s_i] = \frac{|\tau|}{|\tau'| + D(s_i, \tau)} \tag{17}$$

where $D(s_i, \tau)$ is the minimum hops from $s_i$ to all segments on the trajectory $\tau$, measured by the number of road segments needed. $|\tau|, |\tau'|$ is the length of $\tau$ and $\tau'$ and generally $|\tau'| \geqslant |\tau|$, also measured by the number of road segments. Compared with the general geodistance such as Euclidean distance, our RS-T distance considers the alternative routes by introducing $\tau'$, and the form of ratio between the original route and a detoured route also follows the design of *Detour Score* in spatial topological theories [3].

A higher $\mathbf{w}_\tau[s_i]$ means $s_i$ will be more likely to be a positive sample on the alternative paths, with a smaller travel distance or smaller deviation from the actual route. To avoid noise, probabilities in $\mathbf{w}_\tau$ lower than a specific threshold are set to 0. Essentially, this re-weight introduces uncertainty into positive sample selection, such that instead of a "hard" selection, the positive samples are choose with a "soft" weight, contributing to a more robust representation.

Regarding the computational complexity for the RS-T distance, it is worth mentioning that RS-T distance is calculated in the pre-processing, where Dijkstra's algorithm [7] is adopted to calculate the shortest path, with time complexity as $O(|S|^2 log|S|)$.

We then reformulate $\mathcal{L}_{ST}$ weighted by RS-T distance as:

$$\mathcal{L}_{ST_w} = -\frac{1}{|\mathcal{T}|} \sum_{\tau_j \in \mathcal{T}} \left[ \frac{1}{|S|} \sum_{s_i \in S} \mathbf{w}_{\tau_j}[s_i] \cdot \mathcal{I}(\mathbf{h}_{s_j}, \mathbf{h}_{\tau_j}) \right] \tag{18}$$

The weighted function could recognize the road segments that are close to the trajectory as positive samples and assign weights to them. A regional segment $s_i$ with a higher weight will be more co-trained with trajectory semantics, such that regional dependencies can be better preserved.

#### 4.2.4 The Final Loss Function.
In summary, the overall loss function $\mathcal{L}$ is defined as the weighted sum of $\mathcal{L}_{SS}$, $\mathcal{L}_{TT}$ and $\mathcal{L}_{ST}$ with weights $\lambda_{SS}$, $\lambda_{TT}$ and $\lambda_{ST}$ respectively and $\lambda_{SS} + \lambda_{TT} + \lambda_{ST} = 1$.

$$\mathcal{L} = \lambda_{SS} \cdot \mathcal{L}_{SS} + \lambda_{TT} \cdot \mathcal{L}_{TT} + \lambda_{ST} \cdot \mathcal{L}_{ST_w} \tag{19}$$

$\mathcal{I}$ is chosen as a Jensen-Shannon MI estimator [13, 14] since Jensen-Shannon divergence loss is insensitive and robust to the number of negative samples. The formulation is given as:

$$\mathcal{I}(\mathbf{X}, \mathbf{Y}) := \mathbb{E}_{\mathbb{P}}[-sp(-\mathcal{D}(\mathbf{X}, \mathbf{Y})] - \mathbb{E}_{\mathbb{P} \times \tilde{\mathbb{P}}}[sp(\mathcal{D}(\mathbf{X}', \mathbf{Y})] \tag{20}$$

where $\mathbf{X}$ is an input sample, $\mathbf{X}'$ is a negative input sampled from $\tilde{\mathbb{P}} = \mathbb{P}$, and $sp$ is the soft-plus function. $\mathcal{D}$ is the function scoring the agreement between two inputs, defined as the inner product $\mathcal{D}(\mathbf{X}, \mathbf{Y}) = \mathbf{X} \cdot \mathbf{Y}$ in our method.

## 5 EXPERIMENTS

We implement the proposed framework into two real-world datasets and four traffic-related tasks and compare it with the state-of-art methods. We conduct detailed ablation studies to demonstrate the advantage and necessity of each component used in our model.

### 5.1 Datasets and Preprocessing

The two datasets are released by *GAIA*[1] project, Didi company, with two-month car-hailing trip data from two cities, Xi'an and Chengdu, China, as summarized in Table 1. Both datasets provide GPS records for every trip. We collect road network information of the two cities from *Open Street Map*[2] and apply a map-matching algorithm [39] to align GPS points to road segments. With the map-matching process, trajectories are transformed into sequences of road segments, then we filter out trajectories that contain two or fewer road segments or have a duration of less than one minute. It is worth mentioning that the RS-T distance used in Eq. (17) is also calculated before the training.

#### Table 1: Dataset Summary

|  | Xi'an | Chengdu |
| --- | --- | --- |
| # road segments | 6,161 | 6,632 |
| # edges | 15,779 | 17,038 |
| avg. trajectory length (m) | 5,880 | 5,732 |
| avg. road segments per trip | 31.11 | 30.87 |

### 5.2 Downstream Tasks and Benchmark Methods

We use the similar settings in [6] and conduct four downstream traffic tasks, with two road segment-based tasks, and the other two trajectory-based tasks. To reduce the influence of other factors in evaluation, we only use simple neural network architectures, such as a fully connected layer or multi-layer perceptron (MLP), as the inference model for each downstream task. For baseline methods, we only compare our method to the state-of-the-art road and trajectory representation learning methods as well as graph representation learning methods. Methods that are designed only for specific tasks are ignored in this experiment, because our purpose is to learn robust representations which can be used in various applications; however, task-specific methods usually take more task-dependent information as input and contain specialized model components, which result in unfair comparison.

#### 5.2.1 Overview of Evaluation Metrics.
Before introducing the downstream tasks, all evaluation metrics are summarized as follows.

- **Micro-F1 (Mi-F1) and Macro-F1 (Ma-F1)**: Mi-F1 and Ma-F1 are popular in classification task [6]. Ma-F1 computes the F1 independently for each class and then takes the average (hence treating all classes equally), whereas Mi-F1 aggregates the contributions of all classes to compute the average F1.
- **Mean Rank (MR) and Hit Ratio@$K$ (HR@$K$)**: MR and HR@$K$ instead are popular in ranking and recommendation task. MR $\in [1, \infty)$ computes the arithmetic mean over all individual ranks. The top $K$ hit ratio HR@$K \in [0, 1]$ is the fraction of the correct answer being included in the ranking list of length $K$. In this experiment setting, we choose $K = 10$.

[1]https://outreach.didichuxing.com/appEn-vue/dataList
[2]https://www.openstreetmap.org/

**Table 2: Performance comparison on road segment-based tasks.**

| Task | Road Label Classification | | | | Traffic Speed Inference | | | |
|------|---------|---------|---------|---------|---------|---------|---------|---------|
| | Chengdu | | Xi'an | | Chengdu | | Xi'an | |
| | Mi-F1 ↑ | Ma-F1 ↑ | Mi-F1 ↑ | Ma-F1 ↑ | MAE ↓ | RMSE ↓ | MAE ↓ | RMSE ↓ |
| Node2Vec | 0.524 | 0.495 | 0.586 | 0.559 | 7.12 | 9.00 | 6.41 | 8.22 |
| DGI | 0.463 | 0.337 | 0.475 | 0.358 | 6.43 | 8.41 | 6.12 | 7.98 |
| RFN | 0.516 | 0.484 | 0.577 | 0.570 | 6.89 | 8.77 | 6.57 | 8.43 |
| IRN2Vec | 0.497 | 0.458 | 0.531 | 0.506 | 6.52 | 8.52 | 6.60 | 8.59 |
| HRNR | 0.541 | 0.527 | 0.631 | 0.609 | 7.03 | 8.82 | 6.52 | 8.45 |
| Toast | 0.602 | 0.599 | 0.692 | 0.659 | 5.95 | 7.70 | <u>5.71</u> | <u>7.44</u> |
| **Proposed** | **0.637** | **0.629** | **0.729** | **0.701** | **4.69** | **6.85** | **5.02** | **7.08** |

- **Root Mean Squared Error (RMSE) and Mean Absolute Error (MAE)**: RMSE and MAE are metrics used to evaluate a Regression Model. The calculations are as follows: $MAE = \sum |y_i - \hat{y_i}|/n$, $RMSE = \sqrt{\sum (y_i - \hat{y_i})^2/n}$, where $y_i, \hat{y_i}$ are true value and predicted value for the $i$th sample respectively, $n$ is the number of samples.

### 5.2.2 *Road Segment-based Tasks*. 
We consider two typical tasks to evaluate road network representation: (1) road label classification and (2) traffic speed inference.

- **Road label Classification:** This task is similar to the node classification task on graph data. We collect road type labels (*e.g.*, motorway, living street) from *Open Street Map*. Five frequent types of labels are used as prediction target. A classifier with a fully connected layer and a softmax layer is applied directly on the road representations. Mi-F1 and Ma-F1 are calculated to measure the classification accuracy.
- **Traffic Speed Inference:** The inference task uses the average speed on every road segment as the inference objective, which is a regression task. The average speed is calculated with the trajectory data. We apply a linear regression model on the road representations for the inference and evaluate the results with MAE and RMSE.

We compare with following state-of-the-art road network representation methods and graph representation methods:

- **Node2Vec** [11]: It learns $d$-dimensional node vectors by capturing node pairs within $w$-hop neighborhoods via parameterized random walks in the network.
- **DGI** [33]: The classic contrastive learning method applied in graph representation learning. It learns node representations by maximizing the MI between node representation and graph representation.
- **RFN** [15]: It learns representations based on node-relational and edge-relational views of road network graphs, where message passing and interaction are performed.
- **IRN2Vec** [35]: It encodes various relationships between road segment pairs to model road networks. The training pairs are sampled from the shortest paths on the road network.
- **HRNR** [38]: It constructs a three-level neural architecture corresponding to functional zones, structural regions, and

road segments to capture structural and functional characteristics. The model is based on hierarchical GNN.
- **Toast** [6]: It utilizes auxiliary traffic context information to train a skip-gram model. Furthermore, a trajectory-enhanced Transformer module is utilized on trajectory data to extract traveling semantics on road networks.

### 5.2.3 *Trajectory-based Tasks*. 
We choose the following two tasks for trajectory representation evaluation: (1) trajectory similarity search and (2) travel time estimation.

- **Trajectory Similarity Search:** This task aims to find the most similar trajectory for the query trajectory in the database. We use the trajectory representation vectors to compute similarity scores and rank the results in descending order. We take the HR@10 and MR as evaluation metrics.
- **Travel Time Estimation:** Travel time estimation is a regression task to predict the travel time of a given trajectory. We use a 3-layer MLP model as a regression model and evaluate the result with MAE and RMSE.

The following trajectory representation benchmarks are chosen:

- **ParaVec** [17]: It is an embedding method to learn paragraph representations. We treat a trajectory as a paragraph and derive its representation.
- **T2Vec** [19]: It is an encoder-decoder framework trained to reconstruct the original trajectory using noisy and sparse road sequences as inputs. The backbone of both the encoder and decoder are LSTM units.
- **Toast** [6]: It first utilizes auxiliary traffic context information to learn road segment representations, then applies stacked transformer encoder layers to train trajectory representation with route recovery and trajectory discrimination tasks.
- **GTS** [12]: A two-step method that first learns the embedding of each POI in the road network and then encodes the trajectory embedding with a GNN-LSTM network.

## 5.3 Experiment Settings
To train our framework, we adopt a graph encoder based on a two-layer GAT, as well as a sequence encoder stacked from four Transformer encode layers with four attention heads. The training set contains 500,000 trajectories, and the model is trained with an Adam optimizer [30] with a batch size of 256 for 10 epochs.

**Table 3: Performance comparison on trajectory-based tasks.**

| Task | Similar Trajectory Search | | | | Travel Time Estimation | | | |
|---|---|---|---|---|---|---|---|---|
| | Chengdu | | Xi'an | | Chengdu | | Xi'an | |
| | MR ↓ | HR@10 ↑ | MR ↓ | HR@10 ↑ | MAE ↓ | RMSE ↓ | MAE ↓ | RMSE ↓ |
| Para2Vec | 216 | 0.251 | 279 | 0.205 | 220.5 | 302.7 | 244.7 | 345.4 |
| T2Vec | 46.1 | 0.781 | 38.6 | 0.806 | 165.2 | 240.7 | 207.5 | 311.0 |
| Toast | <u>10.1</u> | 0.885 | 13.7 | <u>0.905</u> | 127.8 | 190.8 | <u>175.6</u> | <u>265.0</u> |
| GTS | 11.0 | <u>0.889</u> | <u>12.9</u> | 0.896 | <u>126.3</u> | <u>186.7</u> | 176.1 | 267.9 |
| **Proposed** | **8.87** | **0.928** | **9.54** | **0.912** | **121.9** | **179.5** | **163.6** | **243.5** |

We first obtain representation vectors of road segments and trajectories from benchmark models and the proposed model. Then we input the vectors into various downstream tasks. The dimension of representation vectors is all set to 128.

The trajectory dataset is split into two parts by order of date, one for representation learning and the other for evaluation, and there is no overlap between them.

Road classification and speed inference are evaluated with 5-fold cross-validation. Travel time estimation is trained on a dataset containing 80,000 trajectories and evaluated on another 20,000 trajectories. For trajectory similarity search, we first generate ground truth for 5,000 trajectories which serve as queries. It is implemented by replacing randomly selected path from the original trajectory with another one; the detour length is about 10% of the whole trajectory. Then we put the generated ground truth with another 100,000 trajectories together as the database for query processing.

## 5.4 Results and Analysis

The experiment results of the four tasks are shown from Table 2 to Table 3, with the best result highlighted in boldface and the second-best underlined. The larger the Mi-F1, Ma-F1 and HR@10 are, the better the model is (denoted as ↑). MAE, RMSE and MR are opposite (denoted as ↓).

For both road segment-based tasks and trajectory-based tasks, we can find those general methods for graph or sequence representation learning, such as Node2vec, DGI and Para2Vec, get poor results. The possible reason is that these methods are designed for common graph and sequence data such as social networks or natural language, they fail to consider the special characteristics of road network and trajectories.

Within scale methods, e.g. IRN2Vec and T2Vec perform better than those general methods in road segment-based tasks and trajectory-based tasks respectively, because they utilize rich context information of road network or trajectories. For example, both Node2Vec and IRN2Vec learn road representations by exploring the road pair relationship, but IRN2Vec samples road pairs from the shortest path and train with traffic-related multi-tasks, while Node2Vec only sample with biased random walk.

Toast and GTS achieve great improvement compared to T2Vec on both travel time estimation task and trajectory retrieval task, and Toast also greatly exceeded all other baseline methods on road segment-based tasks. These results show that methods trying to utilize road network or trajectory information to help the representation learning of each other achieve better performance than those within-scale methods, and prove the importance of modeling the inter-relations between the two.

**Table 4: Ablation studies of four variant models without (w/o) each component of $\mathcal{L}_{SS}, \mathcal{L}_{TT}, \mathcal{L}_{ST_w}$ and $\mathbf{w}_\tau$. ∗∗ denotes the most important component and ∗ denotes the second important.**

| Task | Road Classification | | Trajectory Search | |
|---|---|---|---|---|
| | Mi-F1 ↑ | Ma-F1 ↑ | MR ↓ | HR@10 ↑ |
| **Proposed** | **0.729** | **0.701** | **9.54** | **0.912** |
| w/o $\mathcal{L}_{SS}$ | $0.663^*$ | $0.613^*$ | 9.86 | 0.911 |
| w/o $\mathcal{L}_{TT}$ | 0.707 | 0.667 | $15.3^{**}$ | $0.877^{**}$ |
| w/o $\mathcal{L}_{ST_w}$ | $0.653^{**}$ | $0.605^{**}$ | 10.1 | 0.908 |
| w/o $\mathbf{w}_\tau$ | 0.710 | 0.684 | $11.5^*$ | $0.892^*$ |

However, the two-step paradigms in these methods limit their performance, because the gap between the two steps may cause information loss and error propagation. Our method utilize contrastive learning to overcome the shortages. We build bridge directly between the road network and trajectory by maximizing their mutual information and jointly training their representations in an end-to-end manner. Dedicated designs in augmentation and loss function for road-road, trajectory-trajectory, and road-trajectory contrast make it possible to better capture the knowledge across the road network-trajectory pair on top of preserving within-road network and within-trajectory information. As a result, our proposed method outperforms all baselines methods significantly.

## 5.5 Ablation Studies

In our method, three loss components as well as a weight parameter are used to jointly train the representations of road network and trajectories. To demonstrate the impacts of the loss components in our framework, we design ablations on the following variants: (1) $\mathcal{L}_{SS}$ scores MI between each road segments and its context; (2) $\mathcal{L}_{TT}$ scores MI between similar trajectory pairs; (3) $\mathcal{L}_{ST_w}$ is the weighted road-trajectory contrastive loss formulated in Eq. (18); (4) $\mathbf{w}_\tau$ is the metric for positive sampling and weight assigning. We compare the four variants, each by deleting each different model component. Ablation studies are conducted on the dataset of city Xi'an. We choose road classification to evaluate road network representation and similarity search to evaluate trajectory representation.

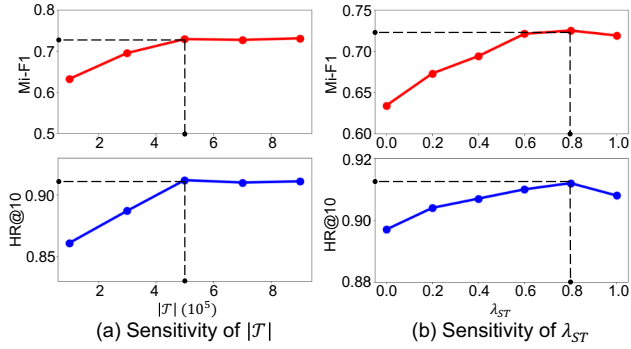

**Figure 3: Parameter sensitivity for (a) the number of trajectories $|\mathcal{T}|$ and (b) road-trajectory contrastive loss weight $\lambda_{ST}$ based on road classification and similarity search task.**

The results are shown in Table 4. The most important component is the one which causes the largest performance loss if discarded. (1) For road segment representation learning and related tasks, the most important component is road-trajectory contrast, and the second comes the road-road contrast. The reason might be because the discrete road segments do not really offer too much information, the trajectories instead bring abundant semantic information to those discrete road segments. (2) For trajectory task, since the trajectories themselves have already enough semantic information, the most important component is trajectory-trajectory contrast. Surprisingly, for the variant w/o $\mathbf{w}_\tau$ is the second most important: when we delete $\mathbf{w}_\tau$, the $\mathcal{L}_{ST_w}$ loss reduces to Eq. (16), which is used in general node-graph cross-scale contrastive learning. Such that, we proved that introducing the RT-S distance weighted loss is better than directly applying node-graph cross-scale contrastive learning. This also means that it is essential to consider the alternative routing information that are commonly happening in real traffic.

Overall, the road-trajectory contrast turns out to have significant effect on both tasks, which proves the component successfully capture the correlation between road network and trajectories.

### 5.6 Parameter Analysis

The most notable difference between our method and previous works is the loss designed for road-trajectory contrast. The distribution of trajectories also influences the performance significantly. Thus, parameter sensitivity is analyzed for the number of trajectories $|\mathcal{T}|$ used in representation learning and road-trajectory contrastive loss weight $\lambda_{ST}$. We use Micro-F1 score in road classification and HR@10 in trajectory similarity search for evaluation.

As shown in Figure 3. (a), training with more trajectories is helpful because it reduces bias. The performance grows linearly first until the growth is capped on a certain amount of trajectories. Thus, there is a trade-off between the performance and training cost. In our setting, to train with 500,000 trajectories is optimal.

As for $\lambda_{ST}$ in Figure 3. (b), the optimal value is 0.8, which means overall $\mathcal{L}_{ST_w}$ is more important than within-scale contrastive loss. Therefore, we believe that the inter-correlation between road and trajectory learned from $\mathcal{L}_{ST_w}$ is generally more essential across various downstream tasks, at the same time $\mathcal{L}_{SS}$ and $\mathcal{L}_{TT}$ are also needed to preserve the within-scale property.

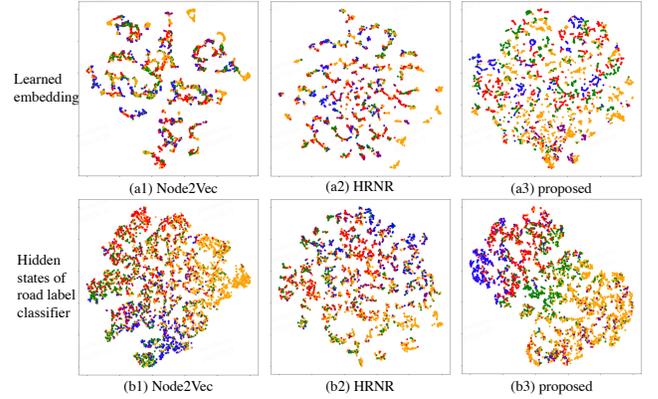

**Figure 4: Embeddings of the road segment representations learned by (a1) Node2Vec, (a2) HRNR and (a3) the proposed method, and hidden states of the road label classifier trained with (b1) Node2Vec, (b2) HRNR and (b3) the proposed method. Different colors mean different road type labels.**

### 5.7 Visualization

To visualize the learned representations, we pick the road label classification task by our model and another two benchmarks, *i.e.*, Node2Vec and HRNR, and visualize them via t-SNE [30] in Figure 4, with (a) visualizing the learned embeddings and (b) demonstrating the hidden states of road label classifier trained with the respective models. From Figure. 4. (a) we could find that our representations are more evenly distributed in the space than Node2Vec and HRNR. This uniformity from our model is the proof of more robust representations to various downstream tasks [36]. We believe this may be because of our joint learning of road networks and trajectories. Figure. 4. (b) shows that in the hidden state of classifier, our method has more clear boundaries between different road types.

### 5.8 Discussion about Generality of the Model

In terms of model generalization, the propose model is a pre-training model, theoretically offering a general representation for road and trajectory. Therefore, we believe it has desired high generality, as demonstrated in its good performance for various downstream tasks. In some scenario where the proposed method needs to be applied to different cities, we do admit there may be some difficulties since different cities have different graphs ($\mathbf{A}_S$ and $\mathbf{A}_T$ in Section 4.2.1). Therefore, the graphs need to be updated and some high-level layers also need to be fine-tuned, but the cost is relatively low, similar to transfer learning.

## 6 CONCLUSION AND FUTURE WORK

In this paper, we studied representation learning on road network and trajectory. Static road network and dynamic trajectory offer semantic and contextual information respectively for representation learning: they are different and complementary. To this end, we propose a unified and end-to-end model that jointly learns road network and trajectory representations by maximizing the mutual information. Precisely, we tailor within-scale contrast for road-road and trajectory-trajectory, and cross-scale contrast for road-trajectory, well-preserving the traffic-domain knowledge. Specifically for road-trajectory cross-contrast, we define a novel distance for road-trajectory to select positive pairs and assign weights adaptively. Via thorough experiments on two real-world car-hailing datasets from two down-steam tasks in road segment and two downsteam tasks in trajectory, we demonstrate the effectiveness and robustness of our method. Via extensive ablation study, we prove the necessity of each component from the proposed model.

In the future, given the temporal information in the trajectory is also useful, we will also introduce the temporal embedding in the trajectory representation.